\documentclass[conference]{IEEEtran}
\IEEEoverridecommandlockouts

\newcommand{\R}{\mathbb{R}}

\newcommand{\F}{\mathcal{F}}

\usepackage{cite}
\usepackage{amsmath,amssymb,amsfonts}
\usepackage{algorithmic}
\usepackage{graphicx}
\usepackage{textcomp}
\usepackage{xcolor}
\def\BibTeX{{\rm B\kern-.05em{\sc i\kern-.025em b}\kern-.08em
    T\kern-.1667em\lower.7ex\hbox{E}\kern-.125emX}}
    
    \newtheorem{theorem}{Theorem}[section]
\newtheorem{lemma}[theorem]{Lemma}

\newtheorem{corollary}[theorem]{Corollary}

\newtheorem{remark}[theorem]{Remark}

\numberwithin{equation}{section}
\begin{document}

\title{Asymptotic convexity of wide and shallow neural networks
\thanks{Work of VSB was supported by a grant from Google Research India}}

\author{\IEEEauthorblockN{Vivek S.\ Borkar}
\IEEEauthorblockA{\textit{Department of Electrical Engineering (retd.)} \\
\textit{Indian Institute of Technology Bombay}\\
Mumbai 400076, India. \\
Email : borkar.vs@gmail.com}
\and
\IEEEauthorblockN{Parthe Pandit}
\IEEEauthorblockA{\textit{Centre for Machine Intelligence and Data Science} \\
\textit{Indiaan Institute of Technology Bombay}\\
Mumbai 400076, India. \\
Email : parthe1292@gmail.com}
}

\maketitle

\begin{abstract}
For a simple model of shallow and wide neural networks, we show that the epigraph of its input-output map as a function of the network parameters approximates epigraph of a convex function in a precise sense. This leads to a plausible explanation of their observed good test performance.
\end{abstract}




\maketitle

\section{Introduction}

There has been considerable interest in analyzing the observed empirical success of wide neural networks, both shallow and deep. A small sample of the enormous activity in this domain can be found in \cite{Cag, Can, Lee, Lu, Mir, Ng, Rad, Wu, Yang}. In this short note, for a simple model of shallow and wide networks, we establish asymptotic convexity for the maps, equivalently their truncated epigraphs (defined below) using a known convexification effect of Minkowski sums of compact sets in $\R^d$. In particular, this suggests that the limiting optimization problem implicit in neural network training for the infinitely wide neural network is a convex minimization problem, therefore amenable to SGD: all local minima are global minima. While the latter property is a consequence of convexity, it does not imply convexity and it may begin to hold for neural networks with `sufficiently large width'. This leads to a plausible explanation of the empirically observed good performance of shallow and wide networks.   

In the next section, we describe the notation used throughout and also state a standard fact from basic real analysis for  later use. The third section recalls the aforementioned result about the Minkowski sum of compact sets. The fourth section describes   its consequences in the present set up.  Section 5 applies this theory to the popular Least Mean Square (LMS) criterion. Section 6 concludes with some remarks.

\section{Notation}

We list here the notation used throughout this article for easy reference.\\

\begin{enumerate}
\item $\F := \{f_\beta : \mathbb{R}^d \to \mathbb{R}^s, \beta \in D\}$ will denote a family of maps $\mathbb{R}^d \to \mathbb{R}^s$  parametrized by a parameter $\beta \in D \subset \R^p$ for some $d,s,p \geq 1$, where $D$ is a compact convex set. We assume that the moduli of continuity of the maps $\beta \mapsto f_\beta(x)$ are uniformly bounded in $x$. \\

\item Let $N \gg 1$ and $f_{\beta_i}, -N \leq i \leq N$, be copies of $f_\beta$ with the subscript $\beta$ distinguished further by the subscript $i$. These represent the input-output maps for our component feedforward neural networks that feed the last layer of the overall shallow and wide network. Let
$$B_N := \{\beta_{-N}, \beta_{-N+1} \cdots , \beta_{-1}, \beta_0, \beta_1, \cdots, \beta_{N-1}, \beta_N\},$$
$$ \qquad 1 \leq N < \infty,$$
$$B_\infty := \{ \cdots , \beta_{-2}, \beta_{-1}, \beta_0, \beta_1, \beta_2, \cdots, \cdots \}.$$

\medskip

\item Let $(X_n,Y^i_n, -N \leq i \leq N), n \geq 1,$ denote the input-output pairs in $\mathbb{R}^d\times\mathbb{R}^s$ associated with the corresponding neural networks. Note that the input is common across the networks. We assume that
\begin{equation}
Y^i_n = f_{\beta_i}(X) + \xi^i_n \label{IOmap}
\end{equation}
where $\{\xi^i_n, n \geq 0,\}$ are i.i.d.\ zero mean noise variables for each $i$, and $\{\xi^i_n, n \geq 0, -N \leq i \leq N; X\}$ is a jointly independent family.\\

\item For a prescribed $\alpha \in (0,1)$, define the `kernel' 
$$K^\alpha_N: \{-N, \cdots , N\} \to [0,1]$$
as
\begin{equation}
K_N(i) := C_N(\alpha)\alpha^{|i|}, \ -N \leq i \leq N, \label{alpha}
\end{equation}
with the normalizing factor $C_N(\alpha)>0$ chosen so as to ensure $\sum_{i=-N}^NK^\alpha_N(i) = 1$. This will serve as the weights for the last layer. That is, the output of the neural network to input $X_n$  at time $n$ is given by
$$\sum_{i=-N}^NK^\alpha_N(i)Y^i_n = \sum_{i=-N}^NK^\alpha_N(i)\left(f_{\beta_i}(X) + \xi_n^i\right).$$

\medskip

\item Let $g(x,y;\beta) := y - f_\beta(x), x \in \mathbb{R}^d, y \in \mathbb{R}^s, \beta \in \mathbb{R}^p,$ denote the error function associated with the above neural networks. For brevity, let $g_i(\beta_i) := g(X^i_n,Y^i_n; \beta_i)$, suppressing the dependence on $n$ because we shall be considering a fixed $n$ for much of our analysis.\\

\item Define $\Phi_N^\alpha: D^{2N+1} \to \mathbb{R}^+, N \geq 1,$ by
\begin{equation}
\Phi^{\alpha,n}_N(B_N) := \sum_{i=-N}^NK^\alpha_N(i)g(X_n, Y^i_n, \beta_i). \label{Phi}
\end{equation}
Note that this is a random function of $B_N$ with $X_n, Y^i_n$ as parameters
\item Likewise, define $K^\alpha_\infty(i) := C(\alpha)\alpha^{|i|}, i \geq 1,$ with $C(\alpha) > 0$ the normalizing factor such that $\sum_{i=-\infty}^\infty C(\alpha)\alpha^{|i|}=1$. Correspondingly, define
\begin{equation}
\Phi^{\alpha,n}_\infty(B_\infty) := \sum_{i=-\infty}^\infty K^\alpha_\infty(i)g(X_n,Y^i_n,\beta_i). \label{Phi1}
\end{equation}

\item Throughout, $Argmin(F)$ for any $F : D \to \R$ will denote the set of global minima of $F$.

\end{enumerate}

We shall also need the following elementary fact from real analysis.\\

\begin{theorem}\label{equic} Let $A \subset \R^d$ be compact and $h_n \in C(A), 1 \leq n \leq \infty$, be equicontinuous. If $h_n \to h$ pointwise, then $h_n \to h$ uniformly. Furthermore, if $x_n \in Argmin (h_n), 1 \leq n < \infty$, then any limit point of $x_n$ as $n \to \infty$ is in $Argmin (h_\infty)$.
\end{theorem}

\medskip

\noindent \textit{Proof sketch:} The first claim is an easy consequence of the Arzela-Ascoli theorem. The second claim follows by letting $n \to \infty$ in the inequality $h_n(x) \geq h_n(x_n) \ \forall x \in D$ along appropriate subsequences and invoking the first claim. \hfill $\Box$

\section{Minkowski sums of truncated epigraphs}

This section recalls a key result about Minkowski sums of compact sets and spells out its implications for truncated epigraphs of functions. 
Recall that the Minkowski sum of sets $A$ and $B$ in a common vector space is defined as $A+B := \{x+y : x\in A,y\in B\}$. Recall also that the \textit{epigraph} of a function $q: G \subset \mathbb{R}^k \to \mathbb{R}$ is defined as $epi(q) := \{(x,y) : x\in G, y \geq q(x)\} \subset \R^{k+1}$. We  take $q$ to be continuous henceforth.\\

 Let $M \gg \sup_{\beta} q(\beta)$. We define the truncated epigraph of $q$, which we  denote by $\widetilde{epi}(q)$, as
$$\widetilde{epi}(q) := \{(\beta,y) : \beta\in D, q(\beta)\leq y \leq M\} \subset \R^{k+1}.$$ 
This will be clearly a compact set by the continuity of $q$ and compactness of $D$. Also define the convex minorant of any $q: D \to \mathbb{R}$, denoted by $q_*$, as the largest convex function dominated pointwise by $q$, i.e., the function
\begin{eqnarray*}
x \mapsto q_*(x)&:=& \max\{h(x) \ | \ h: D \to \mathbb{R} \ \ \mbox{is convex and}\\ 
&& \ \ \ \ \ \ \ h(x) \leq q(x)\}  \ \forall \ x \in D.
\end{eqnarray*}

We have the following result from \cite{Fradelizi}:\\

\begin{theorem} Let $A_n :=$ the $n$-times Minkowski sum of a compact  set $A \in \mathbb{R}^{k+1}$ with itself. Then
$$\frac{1}{n}A_n \ \to \ \overline{co}(A)$$
in the Hausdorff metric, where $\overline{co}(\cdot)$ stands for the closed convex hull. The convergence rate is $O\left(\frac{1}{n}\right)$.\end{theorem}

\medskip

\begin{remark} The survey \cite{Fradelizi} also considers alternative convergence notions. We do not use them here. The result goes back to Starr, in fact in a more general form, motivated by certain problems in economics \cite{Starr, Starr1}. It is based on the celebrated Shapley-Folkman lemma, proved in an unpublished note in response to a query by Starr \cite{Shapley}. 
\end{remark}

\medskip

Let $q$ be as above and $A = \widetilde{epi}(q)$ in what follows.\\

\begin{corollary}\label{mink}  $\frac{A_n}{n}\to\widetilde{epi}(q_*)$ w.r.t.\ the Haussdorff metric.
\end{corollary}

\medskip

\noindent \textit{Proof} From the preceding theorem, we know that $\frac1n A_n \to \overline{co}\left(\widetilde{epi}(q)\right)$. Thus we need to prove that $$ \widetilde{epi}(q_*) =  \overline{co}\left(\widetilde{epi}(q)\right).$$ Since $q \geq q_*$ pointwise and $\widetilde{epi}(q_*)$ is closed and convex, the r.h.s.\ is contained in the l.h.s.
If the two are not equal, there must be a point $x^* \in D$ and $y^* \geq q_*(x^*)$ such that $(x^*,y^*) \notin \overline{co}\left(\widetilde{epi}(q)\right)$. But then there is a separating hyperplane that separates the two. The pointwise maximum of the affine map that defines this hyperplane and $q_*$ would be another convex function lying below $\widetilde{epi}(q)$ and $\geq q_*$ everywhere, with the strict inequality holding at some points in $D$. This contradicts the definition of $q_*$, proving the claim.
\hfill $\Box$

\medskip

Clearly, $Q := Argmin(q) \subset  Argmin(q_*) \subset D$. The former is compact nonempty by continuity of $g$ and compactness of $D$, and so will be the latter, which will necessarily be $\overline{co}(Q)$ by the convexity of $q_*$.
This is immediate from the foregoing. \\

To summarize, Minkowski sums of truncated epigraphs (and therefore epigraphs) tend to the truncated epigraph (resp., epigraph) of the convex minorant. Furthermore, the $Argmin$ of the latter equals the closed convex hull of the $Argmin$ of former. We  connect this with the function $\Phi_N^{\alpha,n}$, defined in \eqref{Phi} in the next section.

\section{Main result} 

We begin with the following lemma.\\

\begin{lemma}\label{Cesaro} $(x,y) \in \widetilde{epi}(q_*)$  if and only if it is a limit of $\frac{1}{n}\sum_{i=1}^n(x_i,y_i)$ for  some $(x_i,y_i) \in \widetilde{epi}(q), \forall\, i \geq 1$. \end{lemma}

\medskip

\noindent \textit{Proof}
The `if' part follows Corollary \ref{mink}.

Conversely, it is clear that $x$ above must be a limit point of $\frac{1}{n}A_n$ as $n\to\infty$. For $\epsilon > 0$ and $m \geq 1$, pick $N(1) = 1$ and $1 \leq N(m)\upuparrows \infty$, $y_i \in A$, $i \geq 1$  such that
$$\left|\frac{\sum_{i= N(m)}^{N(m+1) -1}y_i}{N(m+1)-N(m)}  - x\right| \ < \ \frac{\epsilon}{2^m} \ .$$
This is possible because $x$ is a limit point of  $\frac{1}{n}A_n$ as $n\to\infty$. Then it is easy to check that $\frac{1}{n}\sum_{i=1}^ny_i \to x$.
\hfill $\Box$

\medskip

\begin{remark} In what follows, we shall often deal with two sided sequences 
$\cdots, -x_{n-1}, x_n, x_{n+1}, \cdots$
instead of the usual one sided sequence $y_1, y_2, \cdots$. The former can be mapped to the latter by enumerating it as $z_1, z_2, \cdots,$ with 
$z_1= x_0, z_2 = x_1, z_3=x_2, z_4=x_3, z_5 = x_4, \cdots .$
We shall apply the foregoing and other results stated for one sided sequences to two sided sequences via this mapping.  \end{remark}

\medskip

We now apply the foregoing to $q = g_i$ defined earlier for $-\infty < i < \infty$. We need the following technical fact.  \\

\begin{lemma} The  family $\sum_{i=-N}^NK^\alpha_N(i)g_i(\cdot), N \geq 1$, is bounded and equicontinuous. \end{lemma}

\medskip

\noindent \textit{Proof} This is an easy consequence of our assumption that  the modulus of continuity of $\beta \mapsto f_\beta(x)$ is bounded uniformly in $x$ and therefore so will be  that of its arbitrary convex combinations.  \hfill $\Box$

\medskip

Let $\{q_i\}$ denote copies of $q$ as above.\\

\begin{lemma}\label{Tauber} Suppose the limit 
$$\ell^*(B_\infty) := \lim_{N\to\infty}\frac{1}{2N+1}\sum_{i=-N}^N q_i(\beta_i)$$ 
exists for a prescribed choice of $\{\beta_i\}$, assumed uniformly bounded. Then $$\lim_{\alpha\uparrow 1}\lim_{N\uparrow\infty}\sum_{i=-N}^N K^\alpha_N(i)q(\beta_i) \ \to \ \ell^*(B_\infty)$$
uniformly in the choice of $\{\beta_i\}$. 
\end{lemma}

\noindent \textit{Proof} Using the fact that the $q_i$'s are uniformly bounded, it is easy to check that for $\beta_i \in A, i \geq 1$,
$$\sup_{\{\beta_i\}}\left|\sum_{i=-N}^NK^\alpha_N(i)q_i(\beta_i) - \sum_{i=-\infty}^\infty K^\alpha_\infty(i)q_i(\beta_i)\right| \stackrel{N\uparrow\infty}{\rightarrow} 0,$$
where the uniformity of the convergence follows from the equicontinuity of the family $\sum_{i=-N}^NK^\alpha_N(i)q_i(\cdot), N \geq 1$, proved in the preceding lemma and Theorem \ref{equic}.
Suppose that the limit $\lim_{n\uparrow\infty}\frac{1}{2n+1}\sum_{i=-n}^nq_i(\beta_i)$ exists and is finite.
By a standard Tauberian theorem (see, e.g., Theorem 2 of \cite{Filar}), we have $$\lim_{\alpha\uparrow 1}\left|\lim_{n\uparrow\infty}\frac{1}{2n+1}\sum_{i=-n}^nq(\beta_i) -  \sum_{i=-\infty}^\infty K^\alpha_\infty(i)q_i(\beta_i)\right| =0.$$
Here too the limits are uniform in $\{\beta_i\}$ because of the equicontinuity of the combined family
$$\left\{\sum_{i=-N}^NK^\alpha_N(i)q_i(\cdot), \ \frac{1}{2N+1}\sum_{m=-N}^N q_i(\cdot), N \geq 1\right\}$$
and Theorem \ref{equic}. Combining the two, the claim follows.
\hfill $\Box$

\medskip

Now we apply this to $q = g$.\\

\begin{theorem} Any choice of $\widehat{\beta}^N \in Argmin_{B_N}(\Phi_N^{\alpha,n}), N \geq 1,$ satisfies
\begin{align*}
\lim_{\alpha\uparrow 1}\lim_{N\uparrow\infty}\widehat{\beta}^N \to  Argmin (g_*).
\end{align*}
\end{theorem}

\noindent \textit{Proof} By Theorem 2.4 of \cite{Serfozo}, we have
\begin{eqnarray}
\sum_{i=-N}^NK^\alpha_N(i)g_i(\widehat{\beta}^N_i) &\leq& \sum_{i=-N}^NK^\alpha_N(i)g_i(\beta_i) \nonumber \\
&\to&  \sum_{i=-\infty}^\infty K^\alpha_\infty(i)g_i(\beta_i) \label{one}
\end{eqnarray}
as $N \to \infty$. By Theorem 2 of \cite{Filar}, we also have
\begin{eqnarray*}
&&\liminf_{N\to\infty}\frac{1}{2N+1}\sum_{i=-N}^Ng_i(\beta_i) \\
&\leq&  \liminf_{\alpha\uparrow 1_-}\sum_{i=-\infty}^\infty K^\alpha_\infty(i)g_i(\beta_i) \\
&\leq& \limsup_{\alpha\uparrow 1_-}\sum_{i=-\infty}^\infty K^\alpha_\infty(i)g_i(\beta_i)  \\
&\leq& \limsup_{N\to\infty}\frac{1}{2N+1}\sum_{i=-N}^Ng_i(\beta_i).
\end{eqnarray*}
It follows that whenever $\lim_{N\uparrow\infty}\frac{1}{2N+1}\sum_{i=-N}^Ng(\beta_i)$ exists, we have
\begin{equation}
\lim_{N\to\infty} \ \frac{1}{2N+1}\sum_{i=-N}^Ng_i(\beta_i) =  \lim_{\alpha\uparrow 1_-} \ \sum_{i=-\infty}^\infty K^\alpha_\infty(i)g_i(\beta_i). \label{two}
\end{equation} 
Also, by familiar equicontinuity arguments, this convergence is uniform in $B_\infty$. The claim follows by \eqref{one}, \eqref{two}, and Lemma \ref{Cesaro}.
\hfill $\Box$

Note that uniformity of convergence plays an important role here.

\section{The Least Mean Square criterion}

Here we consider the specific case of the Least Mean Square criterion as an example. Consider  $Y^i_n = f_{\beta^*}(X) +\xi^i_n$ as above. The output of the neural network then is 
$$Y= \sum_{i=-N}^NK_N^\alpha(i)\alpha^iY_n^i$$
and the mean square error is .
\begin{eqnarray*}
\lefteqn{E\Bigg[ \ \Bigg\|\sum_{i=-N}^NK^\alpha_N(i)(Y^i_n-f_{\beta_i}(X_i))\Bigg\|^2 \ \Bigg] } \\
&=& E\Bigg[ \ \Bigg\|\sum_{i=-N}^NK^\alpha_N(i)(Y^i_n-f_{\beta^*}(X_n) + f_{\beta^*}(X_n) - \\
&& \ \ \ \ \ \ \ \ \ \ \ \ \ \ \ \ \ \ f_{\beta_i}(X_n))\Bigg\|^2 \ \Bigg] \\
&=& E\Bigg[ \ \Bigg\|\sum_{i=-N}^NK^\alpha_N(i)(\xi^i_n + f_{\beta^*}(X_n) - f_{\beta_i}(X_n))\Bigg\|^2 \ \Bigg]. 
\end{eqnarray*}
Following our earlier arguments and using the fact that $\xi^i_n, -\infty < i < \infty$, are i.i.d.\ zero mean for each $n$, we have,
\begin{eqnarray*}
\lefteqn{\lim_{\alpha\uparrow 1_-}\lim_{N\uparrow\infty}\left(\sum_{i=-N}^NK^\alpha_N(i)(\xi^i_n + f_{\beta^*}(X_n) - f_{\beta_i}(X_n))\right)} \\
&=& \lim_{N\uparrow\infty}\frac{1}{2N+1}\left(\sum_{i=-N}^N(f_{\beta^*}(X_n) - f_{\beta_i}(X_n))\right)
\end{eqnarray*}
Next, condition on $X_n = x$ and view the right hand side as a function of $B_\infty$.
Once again, the theory developed above tells us that Cesaro sums of the the truncated epigraphs of $(f_{\beta^*}(x) - f_{\cdot}(x))$ will converge in Hausdorff metric to the epigraph of a convex function $\beta \mapsto \tilde{f}_\beta(x)$. However, we are dealing with the map $(x,\beta) \to F_\beta(x) := \beta \mapsto |\tilde{f}_\beta(x)|^2$. There are two possibilities. The first, the less interesting one, is that the map $\beta \mapsto \tilde{f}_\beta(x)$ is non-negative. In this case, its square is also convex and we still have a convex minimization problem. If not, then what we have is a function obtained by flipping the negative part of the graph of a sign-indefinite convex function across the $\beta$-plane. This function will not be convex, but it will nevertheless have all its local minima $=$ global minima along the level set where $F_\beta(x) = 0 = \tilde{f}_\beta(x)$. \\

There is, however, an additional averaging over $\{X_n\}$. This is not accounted for while analyzing the scheme for a fixed $n$ described above, which was carried out `conditioned on $X_n=x$'. To handle this, consider the actual SGD. For simplicity, i.e.,  in order to avoid additional bookkeeping of a routine nature, we consider the `asymptotic' (as $N\uparrow\infty$) SGD given by
$$\beta(n+1) \in \beta(n) - a(n)\nabla^\beta F_{\beta(n)}(X_{n+1}),$$
where $\nabla^\beta$ denotes the subgradient operator in the $\beta$ variable. This is in fact the gradient in $\beta$ when away from the set of minima. The scalars $\{a(n)\}$ are step sizes satisfying the Robbins-Monro conditions
$$a(n) > 0, \ \sum_na(n) = \infty, \ \sum_na(n)^2 < \infty.$$ 
Then
\begin{eqnarray*}
F_{\beta(n+1)}(X_n) &=& F_{\beta(n)}(X_n) -  a(n)\|\nabla^\beta F_{\beta(n)}(X_n)\|^2 \\
&& + \  O(a(n)^2).
\end{eqnarray*}
Since $X_n, n \geq 0,$ are i.i.d., the law of $X_n =$ the law of $X_0$. Using this and taking expectations on both sides, we get
\begin{eqnarray*}
\lefteqn{E\left[F_{\beta(n+1)}(X_0)\right]
 = E\left[F_{\beta(n)}(X_0)\right]} \\
 && - \  a(n)E\left[\|\nabla^\beta   F_{\beta(n)}(X_0)\|^2\right] + O(a(n)^2) \\
&<& E\left[F_{\beta(n)}(X_0)\right] -  O(a(n)^2) 
\end{eqnarray*}
as long as 
$$P(\beta(n) \notin Argmin(F_\cdot(X_n))) > 0.$$  
Since $\sum_na(n)^2 < \infty$, we get convergence of $E\left[F_{\beta(n)}(X_0)\right]$, along with  the conclusion that 
$$ E\left[\|\nabla^\beta F_{\beta(n)}(X_0)\|^2\right] \ \to \ 0.$$
Thus the norm of the gradient approaches zero  in mean square. Hence it also does so in probability. Recall that a sequence of random variables in $\R^d$ converges to a limiting random variable in probability if and only if every subsequence thereof has a further subsequence that converges to the same limiting random variable a.s. Thus consider a subsequence along which 
$$\nabla^\beta\tilde{f}_{\beta(n)}(X_{n+1}) \ \to \ 0 \ \mbox{a.s.}$$
Since 
$$\beta(n) \notin Argmin\left(\tilde{f}_\beta(X_{n+1})\right) \ \Longrightarrow \ 
\|\nabla^\beta\tilde{f}_{\beta(n)}(X_{n+1})\| \neq 0,$$
we must have
\begin{equation}
\inf_{y\in Argmin\left(\tilde{f}_{(\cdot)}(X_{n+1})\right)}\|\beta(n) - y\| \to 0 \label{InProb}
\end{equation}
along this subsequence. Since this holds a.s. for some subsequence of every subsequence, we have \eqref{InProb} hold in probability. 

\begin{remark} Note that we have ignored the local maximum as a candidate limit for the SGD. This is because it is an unstable equilibrium for the SGD and will be avoided a.s.\ under mild conditions on the noise, see \cite{Borkar}, section 3.4 and the references therein. \end{remark}

\end{document}